\documentclass[10pt,twocolumn,letterpaper]{article}
\usepackage{cvpr}
\usepackage{times}
\usepackage{epsfig}
\usepackage{graphicx}
\usepackage{amsmath}
\usepackage{amssymb}
\usepackage{verbatim}
\usepackage{mathtools}
\usepackage{dcolumn}
\usepackage{url}
\usepackage[utf8]{inputenc}

\cvprfinalcopy 

\begin{document}

\title{Empirical Autopsy of Deep Video Captioning Frameworks}
\author{Nayyer Aafaq \hspace{4mm} Naveed Akhtar  \hspace{4mm} Wei Liu \hspace{4mm} Ajmal Mian\\
Computer Science and Software Engineering,\\
The University of Western Australia.\\
{\tt\small nayyer.aafaq@research.uwa.edu.au,} \hspace{-2mm}
{\tt\small \{naveed.akhtar,\hspace{-1mm} wei.liu,\hspace{-1mm} syed.gilani,\hspace{-1mm} ajmal.mian\}@uwa.edu.au}
}

\maketitle

\begin{abstract}
 
 Contemporary deep learning based video captioning follows encoder-decoder framework. In encoder, visual features are extracted with 2D/3D Convolutional Neural Networks (CNNs) and a transformed version of those features is passed to the decoder. The decoder uses word embeddings and a language model to map visual features to natural language captions. Due to its  composite nature, the encoder-decoder pipeline provides the freedom of multiple choices for each of its components, e.g~the choices of CNNs models, feature transformations, word embeddings, and language models etc. Component selection can have drastic effects on the overall video captioning performance. However, current literature is void of any systematic investigation in this regard. This article fills this gap by providing the first thorough empirical analysis of the role that each major component plays in a contemporary video captioning pipeline. We perform extensive experiments by varying the constituent components of the video captioning framework, and quantify the performance gains that are possible by mere  component selection. We use the popular MSVD dataset as the test-bed, and demonstrate that substantial performance gains are possible by careful selection of the constituent components without major changes to the pipeline itself. These results are expected to provide guiding principles for future research in the fast growing direction of video captioning.
 
\end{abstract}

\section{Introduction}
\label{sec:Intro}
Recent years have seen rising research interests in automatic description of images and videos in natural language using deep learning techniques~\cite{aafaq2019video}.
Most recent methods are inspired by the encoder-decoder framework used in machine translation~\cite{zhang2017nonrecurrent, cho2014learning, sutskever2014sequence}. These techniques use Convolutional Neural Networks (CNNs) as encoders to compute fixed/variable-length vector representations  of the input images or videos. A Recurrent Neural Network (RNN), e.g.~vanilla RNN~\cite{elman1990finding}, Gated Recurrent Units (GRU)~\cite{cho2014learning} or Long Short Term Memory (LSTM) networks~\cite{hochreiter1997long} are then used as decoders to generate natural language descriptions. 

\begin{figure*}[t] 
     \centering
     \includegraphics[width=0.8\textwidth]{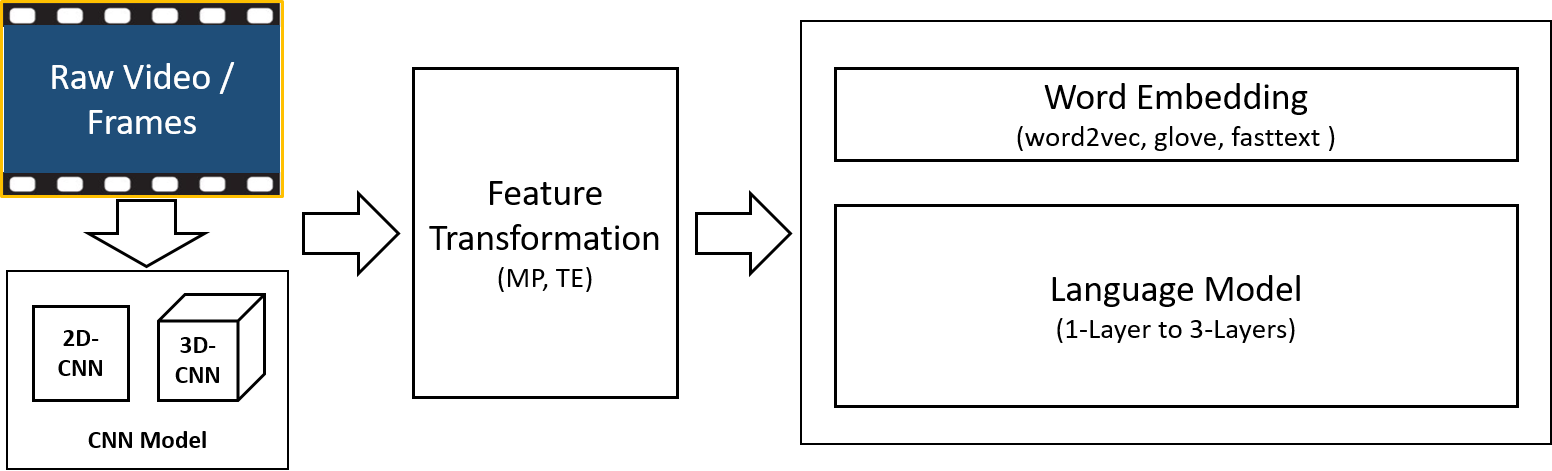} 
     \caption{The encoder-decoder framework of contemporary deep video captioning techniques has four major modules where selections can be made. It is possible to choose from a variety of 2D/3D-CNNs to encode visual features of videos. These features are then transformed to feed to the language model, which can be done by temporal encoding or mean pooling of the features. Multiple choices are also  available for selecting the word embeddings that map words in a vocabulary to dense vector representations to be used by the language model. Language models can have different complexity, governed by e.g.~the number of network layers. In this paper, we vary the choices for each of the four major components, and analyze its effects on the overall captioning performance of the framework.}
\label{fig:captioningframework}
\end{figure*}


In the encoder-decoder pipeline,  visual features are extracted with  2D/3D CNNs from the input videos. These features are then transformed  through  {Mean Pooling (MP)}~\cite{venugopalan2014translating, pan2016jointly, Pan_2017_CVPR}, {Temporal Encoding (TE)}~\cite{yao2015describing}, and/or {Semantic Attributes Learning (SAL)}~\cite{gan2017semantic, Pan_2017_CVPR} before feeding to a language model for natural language caption generation. Most of the existing captioning methods~\cite{chen2015mind, devlin2015language, xu2016msr, pan2016jointly, rohrbach2014coherent, shin2016beyond, yu2016video} mainly differ from each other in terms of the adopted visual feature extraction (i.e. CNN models),  types of visual feature transformations, language models,  and the word embeddings incorporated in the language models. Despite the differences in methods, these four core components are common to nearly all techniques that follow the encoder-decoder framework for video captioning.

In Fig.~\ref{fig:captioningframework}, we take a modular decomposition approach and depict the  encoder-decoder captioning framework in terms of its four core components. Multiple choices are available to instantiate each component. For instance, one can choose either a 2D-CNN or a 3D-CNN as the \textit{CNN model}. Availability of numerous 2D/3D-CNN models provides further flexibility in the choices of the visual feature extraction component. The choice of models may directly affect the overall performance of the video captioning system.  Similarly, the choice of MP or TE for \textit{Feature Transformation} can also have significant effects on the system performance. This modular view of the encoder-decoder pipeline for video captioning is critical to assess how each component contributes to the caption qualities. 
For many existing methods, it is often unclear whether the performance gain is a result of some novel sophisticated enhancements, or simply due to better component selection. This calls for a systematic investigation to quantify performance gains against various component selections across the pipeline. Such analysis would establish a better understanding of the encoder-decoder framework and in turn help identify the most promising components and their best instances. Moreover, it can guide future research in video captioning to focus more on improving the critical components of the framework.

In this work, we present the first systematic analysis of the encoder-decoder framework components with the  aim of revealing the contribution of each component on the quality of the generated captions.
Our analysis is performed by studying the effects of popular choices for each component while keeping the remaining components fixed. We also include the choices of important hyper-parameters in our analysis. The main contributions of this paper are as follows:


\begin{enumerate}
    \item We analyze the visual features of five state-of-the-art CNN models for video captioning under an encoder-decoder framework. We demonstrate that the choice of  CNN model plays a key role in the quality of generated captions. 
    \item We analyze the effects of two popular feature transformations, i.e. mean pooling and temporal encoding, on caption quality. It is observed that temporal encoding can improve performance significantly as compared to the more popular mean pooling transformation.
    \item We explore the affects of popular word embeddings used in language models and report that \emph{FastText}  outperforms the currently  more popular embeddings e.g. \emph{Word2Vec}~\cite{mikolov2013distributed} or \emph{GloVe}~\cite{pennington2014glove}
    \item Lastly, we analyze the effects of language model depth and various hyper-parameters choices  e.g.~internal state size,  number of processed frames, fine tuning word embeddings, and dropout regularization etc.
\end{enumerate}

\section{Setup for Analysis}

We first introduce the setup used in our empirical analysis of the video captioning framework.
For evaluation, we divide the framework into four core components, namely \textit{CNN model} - that encodes visual features of videos, \textit{feature transformation} - that transforms visual features to be used as inputs by the language model component, \textit{word embeddings} - that provides numerical representation of words in the  vocabulary, and the \textit{language model} component, which decodes the visual features into natural language descriptions. Extensive experiments are carried out by varying the methods for each component of the framework and analyze the captioning performance of the overall pipeline. 

We measure the performance in terms of most commonly used evaluation metrics in the  contemporary captioning literature, namely Bilingual Evaluation Understudy (BLEU)~\cite{papineni2002bleu}, Recall Oriented Understudy for Gisting Evaluation (ROUGE)~\cite{lin2004rouge}, Metric for Evaluation of Translation with Explicit Ordering (METEOR)~\cite{lavie2005meteor}, and Consensus based Image Description Evaluation (CIDEr)~\cite{vedantam2015cider}.
These metrics are known to comprehensively evaluate the quality of automatically generated captions. For instance, BLEU measures n--gram based exact matches of the words as well as their order in the reference and generated sentences. ROUGE$_L$ computes the recall score of the generated sentences using n-–grams. METEOR addresses many of the BLEU shortcomings. For example, instead of exact word match, it incorporates synonym matching and performs semantic matching. It is also found to be more robust and closely correlated to human judgments~\cite{kilickaya2016re}. Lastly, CIDEr$_D$ evaluates the consensus between a generated and reference sentences. It has been found to be more robust to distractions e.g.~scene or person changes~\cite{kilickaya2016re}. 

We perform experiments on the popular video captioning dataset MSVD~\cite{chen2011collecting}. This dataset comprises $1,970$ YouTube short video clips,  primarily containing single action/event in a video. Each clip duration varies from $10$ to $25$ seconds. Each video is associated with multiple human annotated captions.  On average, there are $41$ captions per video clip. For bench-marking, we follow the data split of $1,200$, $100$, and $670$ videos for training, validation and testing respectively. This is a widely employed protocol for evaluation using MSVD dataset~\cite{yao2015describing, wang2018m3, gan2017semantic}. We used the Microsoft COCO server~\cite{chen2015microsoft} to compute our results. To clearly analyze the contribution of each component in the overall pipeline, we follow the strategy of freezing all other components when evaluating a particular module. 



\section{Analysis of Framework  Components}
\label{sec:component}
\subsection{CNN selection}
Convolutional Neural Networks (CNNs)  can be readily applied to images and videos. 
In deep learning based encoder-decoder framework for captioning, CNNs dominate the encoder part. 
Due to the significance of a decoder role, the choice of CNN models can affect the overall captioning performance significantly. Hence, we first analyze the five most commonly used CNN models in captioning,  namely;  C3D~\cite{tran2015learning}, VGG-16~\cite{simonyan2014very}, VGG-19~\cite{simonyan2014very}, Inception-v3~\cite{szegedy2016rethinking}, and InceptionResNet-v2~\cite{szegedy2017inception}. 
Among these models, C3D - a popular example of 3D-CNN, is a common choice~\cite{krishna2017dense, yu2016video}
because it can not only process individual frames, but also short video clips.
This is possible due to its ability to process tensors with an extra time dimension. 

\begin{figure}[t] 
     \centering
     \includegraphics[width=\columnwidth]{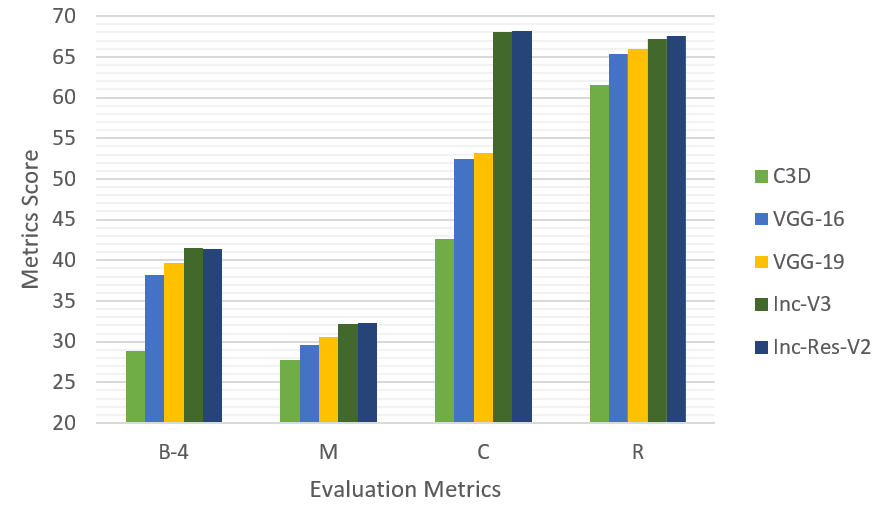} 
     \caption{Performance of five 2D/3D CNNs architectures (C3D, VGG-16, VGG-19, Inception-V3 (Inv-V3), and Inception-ResNet-V2 (Inv-Res-V2)) when used as visual encoder in the captioning framework. Results are  achieved by using Mean Pooling for feature transformation, word2vec as word embedding, and a 2-layer GRU as the language model.}
\label{fig:cnns_mp_dia}
\end{figure}

While performing these experiments to compare different CNN models, we fix all components of the pipeline, except the visual features. For the remaining components, the popular Mean Pooling is used to transform the extracted visual feature into a fixed length vector to represent a complete video;  word2vec word embeddings are used for a 2-layer GRU as the language model.  
The results of this set of experiments are summarized in Fig~\ref{fig:cnns_mp_dia}. We can see a significant variation in captioning performance due to changes in the CNN models, ascertaining that better visual features (obtained from more sophisticated models) lead to better video captions. Hence, a careful selection of CNN model is critical for effective video captioning.  
Interestingly, the spatial visual features of 2D-CNN (\emph{VGG16, VGG19, Inc-V3,} and \emph{Inc-Res-V2}) are able to outperform the spatio-temporal features of \emph{C3D} for the video captioning task, indicating that the extra dimension of 3D CNNs may not be particularly effective in this case.



\begin{figure}[htbp] 
     \centering
     \includegraphics[width=0.8\columnwidth]{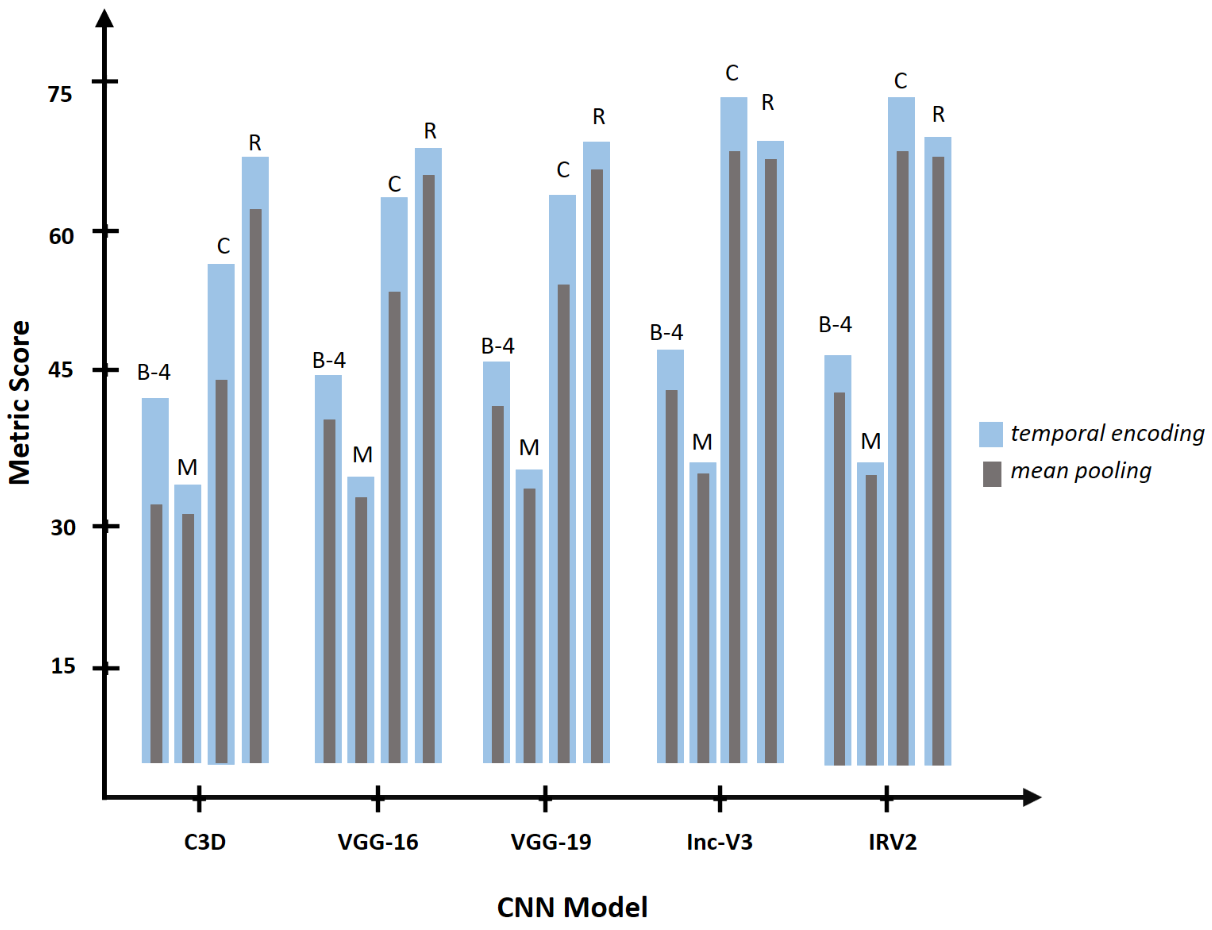} 
     \caption{Performance comparison of five 2D/3D CNNs models with two types of feature transformations i.e. Mean Pooling and Temporally Encoding. It is evident that under all circumstances, temporally encoded features outperform the mean pooled features across all networks and among all metrics.}
\label{fig:cnns_mp_vs_te_dia}
\end{figure}

\subsection{Features Transformation}
\label{sec:FT}
Most existing video captioning methods~\cite{venugopalan2014translating, pan2016jointly, gan2017semantic, Pan_2017_CVPR} perform mean pooling to combine individual frame features into a feature vector for the whole video.  However, this practice is bound to inferior performance as mean pooling can result  in significant loss of temporal information and the order of events in  videos. Such information often plays a crucial role in video understanding for humans. Inspired by this observation, we conduct another series of experiments that compares a temporal encoding strategy to the mean pooling strategy for feature transformation in video captioning. 

For temporal encoding, we follow our previous work~\cite{aafaqspatiotemporal} and compute Short Fourier Transform~\cite{oppenheim1999discrete} of the frame level features of the video. These features are combined in a hierarchical manner that captures the local, intermediate and high level temporal dynamics in the video. Interested readers are referred to our work~\cite{aafaqspatiotemporal} for exact details of the temporal feature transformation. The core insight relevant to our analysis here is that instead of compromising on the temporal information through mean pooling, we capture high fidelity temporal dynamics over whole videos with our temporal encoding. 
Similar to the other sections of this paper, we fix all the remaining components of the underlying framework when analyzing the transformation strategy.
The results of our experiments are summarized in 
Fig~\ref{fig:cnns_mp_vs_te_dia}. It is evident that the models employing temporal encoded features, have outperformed all of their mean pooling based counterparts across all evaluation metrics. Considering the widespread use of mean pooling as the feature transformation strategy in video description, these results are significant. This experiment clearly establishes the supremacy of temporal encoding over mean pooling for the encoders in video description. This temporal encoding does not require any training, as it is applied after the features are extracted, 
therefore little computational overhead is introduced.

It is also evident from the results that for temporal encoding, the performance of different models show a similar behavior relative to each other, which is also the case for the mean pooled features.
For instance, with temporally encoded features, the best performing architecture still remains the best and vice versa is also true. The temporal encoding is providing a significant positive offset to the performance. 

\subsection{Word Embeddings in Language Model}

In this encoder-decoder framework, a word embedding is a vector representation for each word in the available vocabulary for video caption generation. Word embeddings are much more powerful low-dimensional representations for words as compared to the sparse one-hot vectors. 
More importantly, unlike one-hot vectors, word embeddings can be learned for the captioning tasks. 
In captioning literature, two methods are commonly used to compute these vectors.  
The first approach is to learn the vectors from the training dataset while the language  model is trained. In this case, one can initialize the embedding vectors randomly and compute the embeddings tailored to the captioning task. However, such vectors often fail to capture rich semantics due to the  fact that captioning corpus size is often small for the purpose of training a language model. 
The second way to obtain these vectors is to use pre-trained embeddings that are learned for a different task and select those according to the vocabulary of the current task. 

\begin{figure}[t] 
     \centering
     \includegraphics[width=\columnwidth]{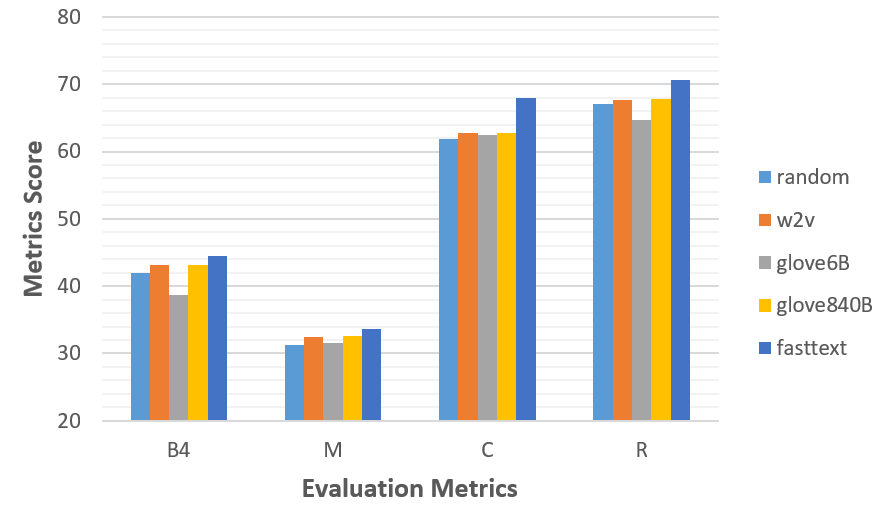} 
     \caption{Performance of four popular pre-trained word embeddings and the learned embedding with random initialization.} 
\label{fig:word_embeddings_dia}
\end{figure}

We follow both of the aforementioned methods to compute embeddings in our analysis.
For the first method, random initialization is performed. For the second, we obtain the most commonly used four  pre-trained word embeddings in the contemporary video description literature, namely  Word2Vec~\cite{mikolov2013distributed}, two variants of  Glove (i.e.~\texttt{glove6B} and \texttt{glove840B})~\cite{pennington2014glove}  and FastText~\cite{bojanowski2016enriching}. 
In our analysis, we select  the best performing CNN model from our experiments in Section~\ref{sec:FT} that uses temporal encoding for feature transformation, i.e.~Inception-ResNet-V2. The results of our experiments for word embeddings are summarized in  Fig~\ref{fig:word_embeddings_dia}. 
From the figure, we can conclude that FastText currently provides much more effective word embeddings for video captioning than the other techniques. Moreover, learning the embeddings with random initialization is still a useful option for the MSVD dataset. This is true to the extent that \texttt{glove6B} shows comparable performance to our randomly initialized learned word embeddings.



\section{Experimental Evaluation}
\label{sec:Exp}

\subsection{Datasets}
We evaluate our technique using two popular benchmark datasets from the existing literature in video description, namely Microsoft Video Description (MSVD) dataset~\cite{chen2011collecting}, and MSR-Video To Text (MSR-VTT) dataset~\cite{xu2016msr}.  We first give details of these datasets and their processing performed in this work, before discussing the experimental results.  

\vspace{3mm}
\noindent \textbf{MSVD Dataset}~\cite{chen2011collecting}: This dataset is composed of  1,970 YouTube open domain videos that predominantly show only a single activity each. Generally, each clip is spanning  over 10 to 25 seconds. The dataset provides multilingual human annotated sentences as captions for the videos. We experiment with the captions in English. On average, 41 ground truth captions can be associated with a single video. For benchmarking, we follow the common data split of 1,200 training samples, 100 samples for validation and 670 videos for testing~\cite{yao2015describing, wang2018m3, gan2017semantic}. 

\vspace{3mm}
\noindent\textbf{MSR-VTT Dataset}~\cite{xu2016msr}:
This recently introduced open domain videos dataset contains a wide variety of videos for the captioning task. It consists of 7,180 videos that are transformed into 10,000 clips. The clips are grouped into 20 different categories. Following the common settings~\cite{xu2016msr}, we divide the 10,000 clips into 6,513 samples for training, 497 samples for validation and the remaining 2,990 clips for testing. Each video is described by 20 single sentence annotations by Amazon Mechanical Turk  (AMT) workers. This is one of the largest clips-sentence pair dataset available for the video captioning task, which is the main  reason of choosing this dataset for benchmarking our technique. 


We attribute the superior performance of FastText to its ability to generate vectors for out of vocabulary words according to the  contextual vectors. The other word embeddings do not have this property. For instance, with $9,914$ words of corpus vocabulary size in FastText, $8,846$ tokens are extracted from the pre-trained embeddings and the embeddings for the remaining $1,068$ tokens are generated using character n-grams of out-of-vocabulary words.
The resulting vectors are then merged to produce the final embedding vector. This strategy is certainly better than random initialization of the out-of-vocabulary words. With FastText at the top,  \texttt{glove840B} and Word2Vec performs almost at par. Among all the pre-trained embeddings, \texttt{glove6B} proved to be the weakest. 



\subsection{Language Model Depth Selection}
In language models, given the type and size of data, depth of the model plays the pivotal role in  effective learning. Where lower layers of a model learn to represent the \emph{syntactic information} (parts of speech, grammatical role of words in each sentence etc.), \emph{semantic information} (meaning of the words, contextual information) is better captured at the higher layers. As each layer learns different type of information, depth of models  becomes important for effective language modelling. However, the modelling performance may start to deteriorate at a certain depth due to the data size limitation.

In our  experiments, Gated Recurrent Units (GRUs) based language models are used. Long Short Term Memory (LSTM) networks are also  popular for language modeling, however, there is a consensus in the literature that the performance and behavior of the two do not deviate significantly for the same task~\cite{chung2014empirical, jozefowicz2015empirical}. 
In our analysis, we vary the number of layers in the language model to observe the performance change. 
Our empirical evaluation shows that two-layers language model performs best under our settings (type and size of dataset, encoder-decoder framework), as compared to a one or a three layer model. The results are summarized in Table~\ref{tab:layers_depth}. Increasing layers from one to {two} generally improves the captioning performance. However, increasing the layers further to {three} does not result in performance gain. In fact, it slightly deteriorates the scores across all metrics. The experiments are performed using visual features of Inception-ResNet-V2 that are transformed with temporal encoding~\cite{aafaqspatiotemporal} for the language model. 


\begin{table}[t]
  \centering
  \small
  \caption{Results of depth variation in GRU-based language model.}
    \begin{tabular}{|c|c|c|c|c|}
    \hline
    Model Depth   & B-4 & M & C & R \\
    \hline
    1 layer     & 49.6 & 34.9 & 75.8 & 71.3   \\
    \hline
    2 layers      & 47.9 & 35.0 & 78.1 & 71.5 \\
    \hline
    3 layers      & 47.7 & 34.6 & 77.4 & 70.8 \\
    \hline
    \end{tabular}%
  \label{tab:layers_depth}%
\end{table}%

\subsection{Hyperparameter Settings}
\label{sec:HP}
Appropriate hyper-parameter setting and model fine-tuning are well-known for their role in achieving the improved performance with deep networks.  Here, we provide a study of a few important hyper-parameters relevant to the captioning task under the encoder-decoder framework. The reported results and findings can serve as guidelines for the community for  training effective captioning models.    

\begin{figure}[t] 
     \centering
     \includegraphics[width=0.8\columnwidth]{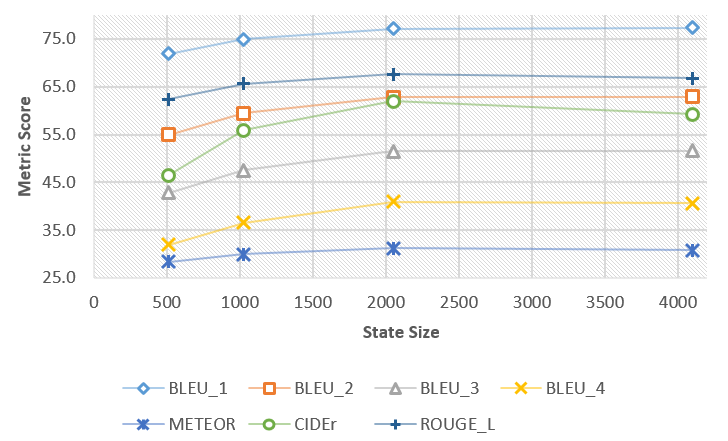} 
     \caption{Performance evaluation of language model using four state sizes. Each trend line show each metric used to compute the captions score.} 
\label{fig:state}
\end{figure}

\vspace{2mm}
\noindent \textbf{State Size}: In the language model, deciding a suitable state size is critical. We tested captioning performance for various state sizes, i.e.~512, 1024, 2048, and 4096.
These results are reported in Table~\ref{tab:state_size_3_layers} and Fig~\ref{fig:state}. We find a direct relation between the state size and the model performance. We compute Pearson's correlation between state sizes and each metric as shown in Table~\ref{tab:state_size_pearson}. It is evident from the results that there is significant correlation between state size and all the metrics. The relationship is even stronger in the lower n--grams of BLEU metric. 
It can be observed that the model performance enhances gradually when we change the state size from 512 until 2048. Further increase in the state size results in  negligible or no improvement in the performance of the BLEU-4, METEOR, CIDEr, and ROUGE$_L$ metrics. However, lower n-grams of BLEU metric (B1, B2, B3) show slight improvements in such cases. 

\begin{table}[t]
  \centering
  \small
     \setlength{\tabcolsep}{5pt} 
  \caption{Results on the state size choices of the GRU language model.}
    \begin{tabular}{|c|c|c|c|c|c|c|c|}
    \hline
    State Size & B1 & B2 & B3 & B4 & M & C & R \\
    \hline
    4096 & 77.3 & 62.9 & 51.6 & 40.7 & 30.8 & 59.2 & 66.7 \\
    \hline
    2048 & 77.1 & 62.8 & 51.5 & 41.0 & 31.3 & 61.9 & 67.6 \\
    \hline
    1024 & 74.9 & 59.4 & 47.5 & 36.6 & 30.0 & 55.9 & 65.6 \\
    \hline
    512 & 71.8 & 54.9 & 42.8 & 32.0 & 28.4 & 46.5 & 62.4 \\
    \hline
    \end{tabular}%
  \label{tab:state_size_3_layers}%
\end{table}%


\vspace{2mm}
\noindent \textbf{Number of Frames}: Frame selection can be treated as a function of time for any video. 
A smaller number of frames reduces the computation cost but at the expense of potentially losing some important spatio-temporal information. In this set of experiments, 
we first process significantly low number of frames, i.e. using every sixteenth ($16^{th}$) frame in the video. In the other experiment, we process all frames of the video. 
A significant gain in model performance is observed in the case when all frames are used.

\begin{table*}[t]
  \centering
  \small
  \caption{Results of Pearson's correlation of state sizes with each metric.}
    \begin{tabular}{|c|c|c|c|c|c|c|c|}
    \hline
     & B1 & B2 & B3 & B4 & M & C & R \\
    \hline
    Pearson's Correlation & 0.8110 & 0.8028 & 0.8107 & 0.7947 & 0.6897 & 0.6635 & 0.6714 \\
    \hline
    p-Value & 0.1889 & 0.1972 & 0.1893 & 0.2053 & 0.3103 & 0.3365 & 0.3286 \\
    \hline
    \end{tabular}%
  \label{tab:state_size_pearson}%
\end{table*}%

\begin{table}[t]
  \centering
  \small
  \caption{Results on how the number of frames affect the captioning quality.}
    \begin{tabular}{|c|c|c|c|c|c|c|c|}
    \hline
    \# Frames & B1 & B2 & B3 & B4 & M & C & R \\
    \hline
    16--F & 65.4 & 46.7 & 34.0 & 23.0 & 24.5 & 31.6 & 57.2 \\
    \hline
    All--F & 69.6 & 52.1 & 39.6 & 28.8 & 27.7 & 42.6 & 61.6 \\
    \hline
    \end{tabular}%
  \label{tab:frames_numbers}%
\end{table}%
These experiments are performed using C3D model \cite{tran2015learning} for visual encoding. The two experiments follows the same settings except in terms of the number of frames used. The results of these experiments are shown in Table~\ref{tab:frames_numbers}. 
As evident from the results, a significant improvement in model performance across all metrics can be observed when all frames are used for captioning. 




\vspace{2mm}
\noindent \textbf{Fine Tuning Pre-Trained Word Embeddings}: It is also observed in our experiments that using pre-trained word embeddings often outperform random initialization based learned embeddings. We also experimented by fine tuning the model  for 10 epochs on the pre-trained word embeddings. It was observed that in this case, the performance on BLEU and CIDEr metrics improved slightly with the fine tuning. However, performance on ROUGE$_L$ metric remained negligible. METEOR metric value showed mixed behaviour with no regular patterns.


\begin{table*}[t]
  \small
  \centering
  \caption{Percentage improvements (Min -- Max) achievable with careful selection of components. First row compared performance including 2D/3D networks. Row 2 demonstrates performance variation with 2D networks only. Subsequent rows show improvements due to word embeddings and the depth of language model.}
    \begin{tabular}{|l|c|c|c|c|}
    \hline
          & B-4 (\%) & M (\%) & C (\%) & R (\%) \\
    \hline
    CNN (2D/3D) & 32.64 -- 44.10 & 6.86 -- 16.61 & 23.0 -- 60.09 & 6.17 -- 9.74 \\
    \hline
    CNN (2D only) & 3.93 -- 8.64 & 3.38 -- 9.12 & 1.53 -- 30.15 & 0.92 -- 3.36 \\
    \hline
    Word Vectors & 8.27 -- 14.99 & 0.64 -- 7.03 & 0.81 -- 9.69 & 0.60 -- 5.37 \\
    \hline
    Depth of Language Model & 0.42 -- 3.98 & 0.87 -- 1.16 & 2.11 -- 3.03 & 0.71 -- 0.99 \\
    \hline
    \end{tabular}%
  \label{tab:analysismisc}%
\end{table*}%


\begin{table}[htbp]
  \small
  \centering
  \caption{Percentage improvement achieved in BLEU (B-4), METEOR (M), CIDEr (C), and ROUGE$_L$ (R) metrics when mean pooled visual features are replaced with temporally encoded features of corresponding networks.}
    \begin{tabular}{|l|c|c|c|c|}
    \hline
          & B-4 (\%) & M (\%) & C (\%) & R (\%) \\
    \hline
    C3D   & 40.97  & 11.91  & 30.75  & 9.42 \\
    \hline
    VGG-16 & 13.09 & 7.77 & 20.04 & 4.59 \\
    \hline
    VGG-19 & 12.59 & 6.86 & 18.80 & 4.70 \\
    \hline
    Inception-V3 & 10.84 & 4.04 & 8.82 & 2.98 \\
    \hline
    Inception-ResNet-V2 & 10.14 & 4.33 & 8.80 & 3.25 \\
    \hline
    \end{tabular}%
  \label{tab:analysis_mp_te}%
\end{table}%
\vspace{2mm}
\noindent \textbf{Dropout in Recurrent Layers}: 
Dropout is a technique used in neural networks to prevent overfitting of the model during training. In recurrent networks e.g.~in GRU, input and recurrent connections to GRU units are probabilistically excluded from activation and weight updates while training the network. Using dropout is typically effective for training language models with large dataset. However, with the MSVD dataset, the caption corpus is rather small ($\sim$48K captions with $\sim$9K unique tokens), dropout therefore does not have a significant effect on language model performance for this dataset, or the datasets of similar scale. We employed dropout in the recurrent layers of language model. 
However, it was observed that application of dropout did not improve the performance. In fact, it sometimes resulted in slight deterioration of the model performance. Based on the observed behavior, we can confidently recommend to avoid the use of recurrent dropout in a GRU language model, given the training data of MSVD size (or comparable) and model complexity of 2-3 GRU layers. 

\vspace{-2mm}
\section{System Level Discussion and Analysis}


With Section~\ref{sec:component} focusing on `ablation study' of individual components, in this section, we further discuss and analyze the results 
of the pipeline as a whole, at the system level.
First, we discuss the results in terms of Min -- Max improvements in captioning score for each metric, respectively, as shown in Table~\ref{tab:analysismisc}. Here, `Min' denotes the minimum percentage gain in the performance which is computed as the difference between the lowest score and the second lowest score in our experiments.
`Max' denotes the percentage gain achieved by comparing the lowest and the highest values achieved. The Min -- Max ranges provide an estimate of the performance gain that is possible by varying the selection of component variants. 

In Table~\ref{tab:analysismisc}, first two rows depict improvements by selecting superior or inferior CNNs (in terms of their original results on ImageNet classification accuracies). These results are obtained using mean pooling strategy over the frame level features of the corresponding networks. When compared across 2D/3D CNNs (first row) we see a drastic obtainable improvement in the model performance i.e.~up to 44 \% in BLEU and 60 \% for CIDEr metric, if we choose the right visual feature encoding model. Similarly, when comparing among 2D CNNs  only (second row), we see there are significant performance variations. These variations only resulted from varying the CNN model. Hence, we can conclusively argue that superior CNNs (with better representation power) can result in significant performance improvement for the captioning techniques. 

The evaluation results for the word vector representations are shown in row 3 of  Table~\ref{tab:analysismisc}. We can observe that there are a few instances of significant performance variations across all metrics when we use different word embeddings. Among the used popular embeddings, \texttt{FastText} performs the best and \texttt{glove6B} the weakest. 
Note that the results also include the learned word vectors obtained during language model training with random initialization.
Moreover, we also experimented with fine tuning of the pre-trained embeddings for 10 epochs for the captioning task. However, we observed that fine tuning does not result in any  drastic performance gain. We noticed that the performance of \texttt{word2vec} and \texttt{glove840B} mostly remain at par with each other. 
Compared to the visual feature encoder selection, we can see the performance gain by the informed selection of word embeddings are not negligible either. 
However, the right CNN model does have a dominant effect on the performance gain as compared to the word embedding selection.  

The last row of Table~\ref{tab:analysismisc} provides language model depth analysis. 
Relative to the gain obtainable by varying other components in the pipeline, altering the depth from \textit{1-layer} to \textit{3-layers}, does not boost the performance significantly. The metric scores generally improve when model depth is varied from \textit{1-to-2} \textit{layers}. However, further increase in the depth degrades the model  performance. We can confidently claim that under the employed popular pipeline, \textit{2-layers} GRU network performs better as compared to the single or three layers RNNs. 


In Table~\ref{tab:analysis_mp_te}, we report the maximum percentage gains caused by the feature transformation techniques in our experiments. Each row reports the metric score improvement resulted when mean pooled features are replaced with the temporal encoding features~\cite{aafaqspatiotemporal} of same CNN network. As can be observed,  there is significant improvement in the model performance across all metrics and all networks with the temporal encoding. The largest  performance gain results in the case of C3D. We conjecture that a major reason behind this phenomenon is that there are always less number of clips as compared to the number of frames in videos. C3D exploits clips which reduces the number of unit data samples containing distinct pieces of information for the task at hands. The temporal encoding strategy is able make up for this discrepancy. Moreover, spatial  feature capturing with a 2D-CNN followed by temporal encoding result in more discriminative video-level features as compared to the  spatio-temporal features of 3D-CNNs. 

Based on the evaluations performed with  all components of the captioning framework, we can order the components in terms of their importance/contribution to the overall captioning performance. To that end, in our experiments, the most significant contribution comes from the feature transformation technique i.e.~Temporal Encoding. The second significant performance variation is possible through the selection of appropriate CNN model. At the third position in terms of contribution to captioning quality, we can place the word embedding vectors. 
The number of network layers in captioning model had less significant role to play in our experimental results. A simple 2-layer GRU seems a reasonable baseline choice for the captioning models. Similar to the network layers, other hyper-parameters choices also contribute to the captioning performance, as mentioned in Sec.~\ref{sec:HP}. However, assuming a reasonable default hyper-parameter settings, their role is far less significant than the variation in the major components of the pipeline. 
\section{Conclusion}
In this paper, we decompose encoder-decoder based automatic video captioning framework into four core components. This allows us to carry out a comprehensive and fair ablation study at both the component level and the system level on a common dataset. 
The  four core components include CNN model (visual feature encoder), feature transformation (e.g.~mean pooling, temporal encoding), word embeddings, and language model. 
Various model hyper-parameters (depth, state size, and dropout in recurrent layers etc.) are also included in our empirical study. 
Exhaustive experiments are carried out for each component to capture the contribution and effects of that component in the overall captioning performance. In particular, 5 popular CNNs (C3D, VGG-16, VGG-19, Inception-V3, and Inc-ResNet-V2), 2 feature transformation algorithms (mean pooling, temporal encoding), 5 Word Embedding Techniques (learned, word2vec, glove6B, glove840B, and fasttext) and an  RNN language model with three different depths (1, 2, 3 layers) are tested. It is found that 
with a well-informed selection of the  components in the encoder-decoder based video captioning framework, a significant performance gain can be achieved. In our experiments, the best performing combination is Inception-ResNet-V2 as the visual encoder, followed by temporal encoding for feature transformation, followed by the use of fasttext word embeddings with a two layer language model.

\section{Acknowledgement}
This work is supported by Australian Research Council Grant ARC DP19010244. The GPU used for this work was donated by NVIDIA Corporation. 
\newpage
{\small
\bibliographystyle{ieee}
\bibliography{refs}
}
\end{document}